\documentclass[runningheads]{llncs}
\usepackage{graphicx}
\usepackage{times}
\usepackage{epsfig}
\usepackage{graphicx}
\usepackage{amsmath}
\usepackage{amssymb}
\usepackage{comment}
\usepackage{float}
\usepackage{subcaption}
\usepackage{caption}

\begin{document}
\title{How Does Heterogeneous Label Noise Impact Generalization in Neural Nets?}
%
%

\author{Bidur Khanal\inst{1}\and
Christopher Kanan\inst{1,2,3}}

\authorrunning{B. Khanal et al.}
%
\institute{Rochester Institute of Technology, USA \\ \email{\{bk9618,kanan\}@rit.edu} \and
Paige, USA \and Cornell Tech, USA}

\maketitle              
\begin{abstract}
   Incorrectly labeled examples, or label noise, is common in real-world computer vision datasets. While the impact of label noise on learning in deep neural networks has been studied in prior work, these studies have exclusively focused on homogeneous label noise, i.e., the degree of label noise is the same across all categories. However, in the real-world, label noise is often heterogeneous, with some categories being affected to a greater extent than others. Here, we address this gap in the literature. We hypothesized that  heterogeneous label noise would only affect the classes that had label noise unless there was transfer from those classes to the classes without label noise. To test this hypothesis, we designed a series of computer vision studies using MNIST, CIFAR-10, CIFAR-100, and MS-COCO where we imposed heterogeneous label noise during the training of multi-class, multi-task, and multi-label systems. Our results provide evidence in support of our hypothesis: label noise only affects the class affected by it unless there is transfer.

\end{abstract}
\section{Introduction}
Supervised deep learning models have been successful in various tasks such as large-scale image classification, object detection, semantic segmentation, and many more \cite{alexnet,fasterrcnn,ronneberger2015unet}.
One of the significant contributions behind the success of supervised deep learning is the availability of well-labeled large datasets. However, such well-labeled datasets are only available for a handful of problems \cite{imagenet_cvpr09,lin2014microsoft}. Often tools like Amazon Mechanical Turk \cite{Buhrmester_2011} and Computer Vision Annotation Tool (CVAT) \cite{openvinotoolkitcvat} are used to label them. The problem with these tools is that they are expensive and require significant time and human effort to label. To circumvent that, many datasets in the real world are either incompletely labeled or extracted from sources that inherently contain label noise \cite{Frenay_2014}.

\begin{figure}[t!]
\centering
\includegraphics[width=0.9\textwidth]{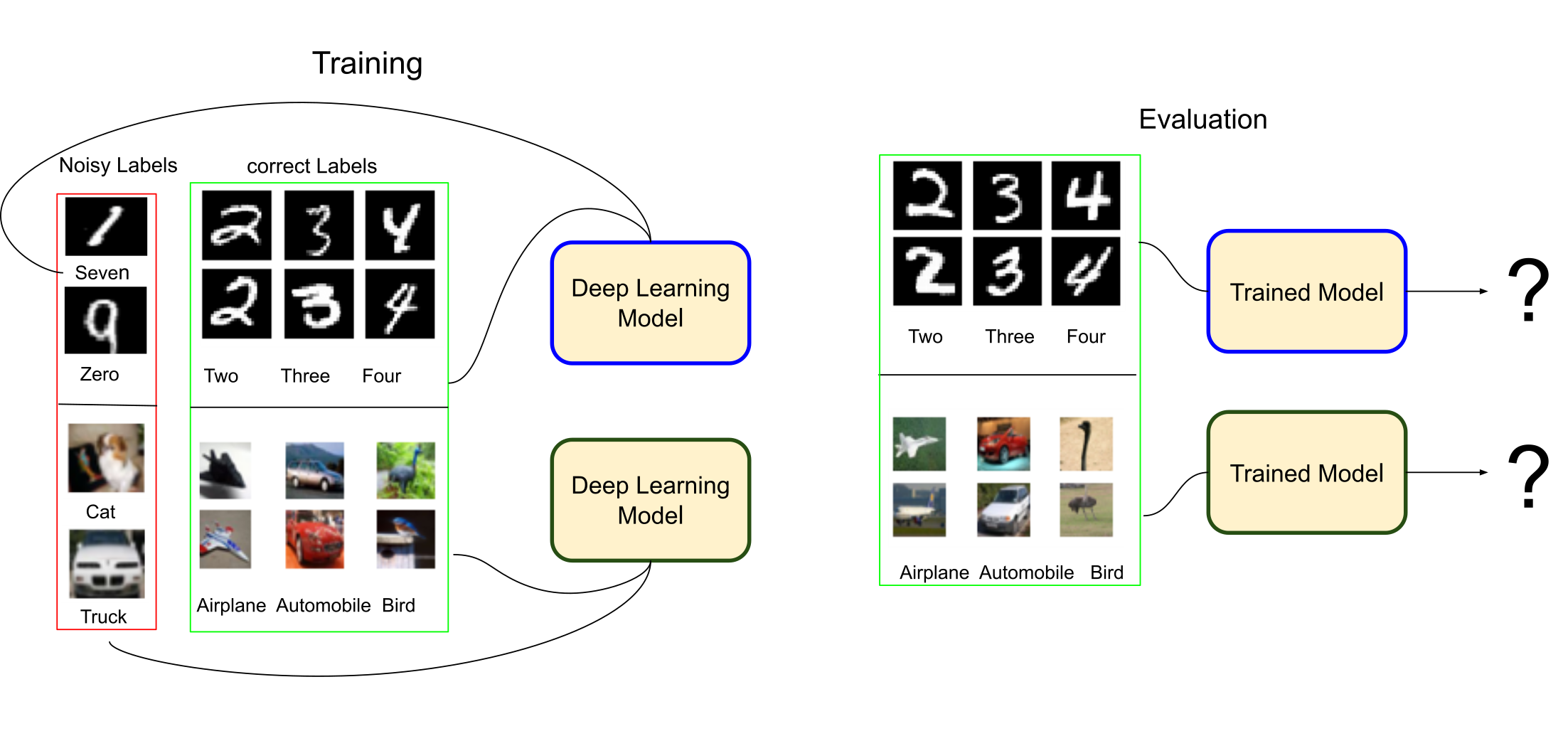}
\caption{ An example of how class-dependent heterogeneous label noise is introduced by corrupting the labels of CIFAR10 and MNIST Dataset. We investigated the impact of noisy labels (red) on the model's performance on clean labels (green).}
\label{fig:heterogeneous}
\end{figure}

Label noise is detrimental to the training of any deep learning model as it directly impacts the model's learning ability \cite{Zhu_2004}. Vision tasks learned with noisy labels don't generalize well, resulting in poor test performance\cite{Zhang_2021}. It is essential to thoroughly study the impact of noisy labels to understand how they are associated with poor performance. This knowledge can be used to improve the current methods that learn with noisy labels \cite{Li_2019}. However, to the best of our knowledge, most of the works studied up to now have mainly focused on examining the performance of the deep learning model under the influence of homogeneous noisy labels imposed by corrupting all the true labels with the same degree \cite{Algan_2021,SANCHEZ_2020}.

We know that the noise may not always be homogeneous and can depend on various heterogeneous sources \cite{zhang2021learning}. Some of the labels might be affected to a greater extent than others because of which the label noise is heterogeneous in nature. The previous studies have not thoroughly investigated the heterogeneous case in supervised vision tasks. Therefore, some open questions still exist. For example: what is the impact of heterogeneous noisy labels of certain classes on the performance of a class with clean labels (as shown in Fig \ref{fig:heterogeneous}) when they are trained together in a naive classification setting? We want to examine to what extent the noise-free class is affected by class-dependent noisy labels. We further extended the question to study the impact in other classification settings, such as multi-task and multi-label learning. 

Multi-task learning\cite{learning} is an approach where a single network is trained to perform two or more tasks. While training a multi-task network, the tasks could positively or negatively interfere resulting in positive or negative transfer respectively. Positive transfer improves the performance of another task, while negative transfer impacts its performance. We hypothesized that if there is a positive transfer between two or more tasks, then training with noisy tasks should impact the performance of clean tasks. The transferred benefit obtained by training tasks together should drop with an increase in label noise in helping tasks. In this work, we verified our hypothesis with experiments. 

Finally, we also investigated the impact of label-dependent noisy labels in multi-label learning, in which noise is present only in a certain group of labels, and examined the impact on the group with clean labels. In summary, the key contributions of our work are:

$\bullet$ Using the popular vision datasets: MNIST, CIFAR10, CIFAR100, and MS-COCO dataset, we assessed the impact of class-dependent, task-dependent, and label-dependent heterogeneous noisy labels on multi-class classification, multi-task learning, and multi-label learning settings, respectively with an attempt to the fill gap that previous studies didn't cover. 

$\bullet$ By investigating task-dependent heterogeneous noisy labels, we showed that if there is a positive transfer from one task to another, inducing label noise in helping task should also impact the performance of other tasks that have clean labels. The drop in task transfer benefit is proportional to the number of noisy labels in helping tasks, i.e., the higher the noisy labels, the higher the transfer drop.

\section{Related Works}
Label Noise is a topic of interest in the deep learning fraternity with a large number of published works. Several surveys provide comprehensive literature reviews on its impact. Song et al. \cite{Song} discussed the generalization problems introduced by label noise in supervised learning methods. They categorically reviewed the state-of-the-art methods used to improve the robustness against label noise. Zhu et al. \cite{Zhu_2004} categorized noise into attribute noise and class noise (label noise) to study their impacts separately and highlighted the class noise to be more harmful. Frenay et al.\cite{Frenay_2014} further investigated the source of label noise and its consequences in learning. Algan et al. \cite{Algan_2021} discussed various noise model-free methods and noise model-based methods to train deep neural networks for the image classification tasks efficiently. These studies highlight the growing interest in developing algorithms that can learn with noisy labels. 

Nowadays, several approaches have been used to improve the existing methods for robust noisy label learning. Some methods used techniques to reduce the influence of incorrect labels in learning \cite{NEURIPS2018_a19744e2}, while some modified the loss function \cite{Wang_2019}. The possible negative influence of noisy labels can be predicted with reliable techniques. Yao et al.\cite{yao2017deep} investigated the quality of the feature embedding that can be used to identify how well the noisy labels could be trusted. Ghosh et al.\cite{ghosh2021contrastive} initialized the network with embeddings learned from contrastive learning to improve the supervised network for classification under noisy conditions. Zhang et al.\cite{Zhang_2020} and Yi et al. \cite{Yi_2019} used the label correction method to improve the model's robustness. While Song et al. \cite{DBLP:journals/corr/abs-1911-08059}, Harutyunyan et al.\cite{harutyunyan2020improving} and Lyu et al. \cite{lyu2020curriculum} focused on improving the model's generalization ability itself. Nishi et al. \cite{nishi2021augmentation}used a combination of weak augmentation and strong augmentation techniques to obtain the best performance in the dataset under noisy labels.

Every year, there is a growing number of papers that uniquely approach this problem. Most of these methods test the robustness of their method by introducing the noise through instance-independent homogeneous corruption of labels in clean data. Lui et al. \cite{liu2021importance} highlight the challenges introduced by instance-dependent label noise and showed that existing methods that learn from noisy labels fail at instance labels. In their analysis, they discussed the problem of memorizing instance-dependent noisy labels. Cheng et al. \cite{cheng2021learning} and Xia et al. \cite{xia2020partdependent} proposed methods to deal with bounded-instance and label-dependent noisy labels. Though we see emerging interest in this area, we haven't found works that thoroughly studied the impact of heterogeneous and dependent noisy labels, as a function of noise strength, in different learning settings. With this works, we have tried to fill the gap in that direction, presenting some insights and verifying our hypothesis that label noise only affects certain classes if there is transfer.

\section{Problem Setup} 
 \label{section:problem_setup}
We designed some methods to systematically corrupt the labels and introduce heterogeneous label noise. We studied in three major classification settings: multi-class classification, multi-task learning, and multi-label learning.

\subsection{Multi-class classification}
 \label{subsection:problem_setup_class}
In a multi-class classification problem, we have inputs $[\bf{X_1}, \bf{X_2},..,\bf{X_N}]$ and our goal is to assign correct target label to each inputs. Let us suppose there are four possible target labels such that an input only corresponds to a particular label $t \in [t_1, t_2, t_3, t_4]$. 
To introduce label noise, we first select a label (say $t_1$), which won't be corrupted (we term this \textit{uncorrupted} or \textit{uncorrupt} label). The other remaining labels $[t_2, t_3, t_4]$ (grouped into a list termed as corrupt list) are corrupted based on a probability value $p$, which determines the strength of corruption, i.e. 0 means that a label will never be changed, while 1 means the label is always replaced with other labels. We maintain a corrupt target list that contains all the possible labels that a corrupted label can take. While corrupting certain labels, we replace them with a randomly chosen value from the corrupt target list. 

There are two possible ways to create a corrupt target list: include the uncorrupt label  $t_1$ along with all other labels or just include labels from the corrupt list. For instance, to corrupt a label $t_2$, we might randomly replace it with a label from a target label list $[t_1, t_3, t_4]$ (\textbf{strategy 1}) or from $[t_3, t_4]$  (\textbf{strategy 2}). We have used both strategies in our experiments and separately studied them. 
At the end of the training, we examined the test classification accuracy in both corrupted and uncorrupted label. Our primary interest is in the uncorrupted label because we want to measure the impact on its performance while training with noisy labels (corrupted).

\subsection{Multi-Task Learning}
In multi-task learning, we aim to boost the performance in some tasks by training tasks together. Let $\text{T}_1$ and $\text{T}_2$ be any arbitrary tasks that a joint network should learn together. In this setup, we first find a task $\text{T}_1$ that benefits by training together with the other task $\text{T}_2$. The overall performance of task $\text{T}_1$, when there is positive transfer, should be higher than that when task $\text{T}_1$ is trained alone. Our hypothesis says that inducing label noise in a helping task, i.e. $\text{T}_2$ should impact the performance of task $\text{T}_1$, resulting in a drop in its test performance. First, task $\text{T}_1$ is trained independently to establish a single task performance baseline. Then, a shared network is trained jointly on tasks $\text{T}_1$ and $\text{T}_2$, such that positive transfer benefit is obtained in task  $\text{T}_1$.

After the transfer benefit is seen in task $\text{T}_1$, the labels of task $\text{T}_2$ are corrupted by randomly changing the labels within the task. The labels of task $\text{T}_1$ shouldn't be changed as we are interested in studying the impact in a clean task on introducing noise in the helping task. The labels are corrupted in a similar manner described in multi-class classification. In both the clean and corrupted tasks, the test performance is measured as a function of label corruption strength in helping tasks. We only used two tasks to make our experiments simpler and tractable.  

\subsection{Multi-Label Learning}
Unlike the multi-class classification problem, in multi-label learning problem, the inputs $[\bf{X_1}, \bf{X_2},..,\bf{X_N}]$ can belong to one or more target labels  $[t_1, t_2,...t_M]$, where N is the total number of input samples and M is the total number of possible labels. The labels are no more treated as mutually exclusive targets, and the input-output mapping can be one-to-many. We are interested in seeing how the label noise in a certain category impacts the prediction performance in a group of clean labels.

Let $L_1$, $L_2$, $L_3$ and $L_4$ be any four arbitrary labels. We can divide the labels into two categories, i.e., uncorrupted and corrupted lists. For instance, the uncorrupted list may contain label $L_1$ and $L_2$, and  corrupted list may contain other two labels $L_3$ and $L_4$. The labels can be grouped in any combination. We corrupted labels in the corrupted list of training data by randomly swapping the labels within the list with some probability as described in previous setups.After training, the test performance of both uncorrupted and corrupted labels is evaluated as a function of corruption strength.

\section{Experiments and Datasets}

We now describe the three categories of experiments designed to study the impact of heterogeneous noisy labels. The datasets and network architecture selection are made based on the requirements and design of the experiments.

\subsection{Multi-Class Classification}
Multi-class classification is a well-studied problem; therefore, there are many open-source datasets available on the web. In our study, we have used two popular datasets: MNIST \cite{lecun-mnisthandwrittendigit-2010} and CIFAR10 \cite{cifar10}. MNIST contains 60,000 training images and 10,000 testing images of handwritten digits. Each image is a gray-scaled image with $28 \times 28$ dimensions and represents one of ten digits. CIFAR10 contains 50000 training and 10000 test RGB images of dimension $32 \times 32$ categorized into ten classes. To classify the MNIST, we have used a basic LeNet-5 CNN architecture with dropout, ReLU activation in the hidden layer, and a final softmax layer. For CIFAR10, we have used a ResNet18 based architecture. 

Using the method discussed in the setup section \ref{subsection:problem_setup_class}, we corrupted any nine out of ten classes in training sets of MNIST and CIFAR10 with a certain probability value while keeping the tenth class as it is. The mean classification accuracy of each corrupted class and the uncorrupted class was evaluated at each corruption probability. We chose eight corruption probability values from 0 to 1 to experiment with, and at each probability, we performed ten sets of experiments, using a different uncorrupted class at each for robustness. There were 80 sets of training experiments combining eight corruption probability and ten sets of experiments. In all of these eighty experiments, we didn't include the uncorrupted class in any of the corrupt target lists. We also performed another 80 sets of experiments with the exact setting, but this time, including the uncorrupted class in the corrupt target lists. For the final result, we averaged the values at each corruption probability.

\subsection{Multi-Task Learning}

We used two tasks to study the impact of task-specific label noise in a multi-task learning scenario. We split 100 classes of the CIFAR100 \cite{cifar100} into two 50-class classification tasks: Task1 and Task2 and trained multi-task network. We monitored the validation loss in Task1 to save the best model for testing.

In our experiment, we tried three random splits for robustness. Each split had different classes assigned to the tasks. After the splits, we also trained a single task network for task $\text{T}_1$ of each three splits, which provided us the baseline to compare multi-task performance. We averaged the single task performance of all the splits to get a mean single-task performance. Similarly, we also averaged the multi-task performance of all the splits across respective tasks and corruption probability to find respective mean values. 


\begin{figure}[t]
\begin{center}
\includegraphics[width=0.6 \textwidth]{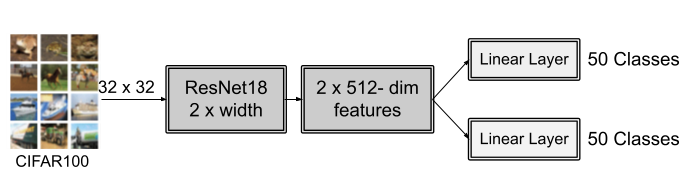}
\caption{{Overview of Multi-Task learning Architecture that was used in our experiment. CIFAR100 was split into two tasks each containing 50 classes. The CNN features of ResNet18 with twice the original width were used as the shared features.
{\label{fig:multitask1}}
}}
\end{center}
\end{figure}

\begin{figure}[t]
\begin{center}
\includegraphics[width=0.9 \textwidth]{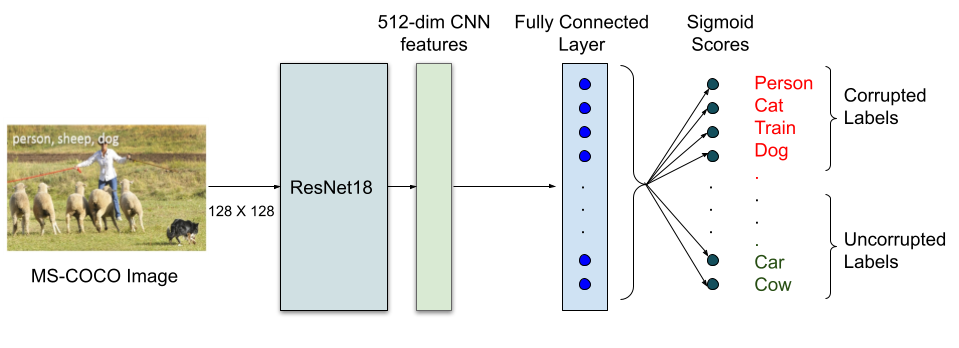}
\caption{{Overview of Multi-Label Classification Architecture that was used in our experiment. The last CNN layer of ResNet18 was connected to an 80-dimensional fully connected layer having sigmoid activation function. We corrupted a group of labels (corrupted labels) by randomly changing their values to other values within the group and trained them with the unchanged (uncorrupted) group of labels.
{\label{fig:multilabel1}}
}}
\end{center}
\end{figure}

As shown in Figure \ref{fig:multitask1}, we used a 2-head classification network with a shared ResNet18 backbone. We doubled the width of the ResNet18 architecture to obtain a higher transfer benefit than its single-task counterpart with the same width. Further, we used Mish activation function \cite{misra2020mish} instead of ReLU because of its higher performance \cite{misra2020mish}. The images from two tasks were fed to the network alternatively while training. The categorical cross-entropy loss was computed separately for each output head and then summed and back-propagated. We then introduced label noise in Task2 by corrupting the labels of 50 classes with some probability as done in multi-class classification. The test classification accuracy in both the tasks at that particular corruption probability was evaluated. For this experiment, we chose 12 different probability values in the range from 0 to 1. At each corruption probability, we trained three times and averaged the test accuracy for robustness.
\subsection{Multi-Label Learning}
 \label{section:multi-label}

For multi-label learning, we used a small subset of MS-COCO dataset; we call it mini MS-COCO \cite{giddyyuppcoco-minitrain}. This version contains randomly selected $20\%$ MS-COCO training images such that datasets' statistics match that of the original dataset. We downsampled all the images to the fixed size of 128 $\times$ 128. An image can contain one or more labels out of 80 labels.

We modified the Resnet18 for multi-label classification by changing the softmax activation function in the last layer into a sigmoid function for each output node. Instead of outputting a probability value for each node, the network outputs a one-hot encoded vector.  We induced label noise in the training examples by corrupting the class labels. We divided the 80 labels into two categories: \textbf{category 1}, which contained corrupted labels and \textbf{category 2} which contained uncorrupted labels. The labels in \textbf{category 1} were corrupted by randomly changing the true labels to other labels within the category using a certain probability, while \textbf{category 2} was not changed at all. Corruption can be done in two ways: either the number of corrupted labels is greater or less than the number of uncorrupted labels. For both the case, we experimented with 12 different probability values in the range from 0 to 1 and measured mean average precision for each.

\section{Results }
We now analyze the results in multi-class classification, multi-task learning and multi-label learning settings.

\begin{figure}[t]
\centering
  \begin{subfigure}{0.49\textwidth}
    \includegraphics[width=1\linewidth]{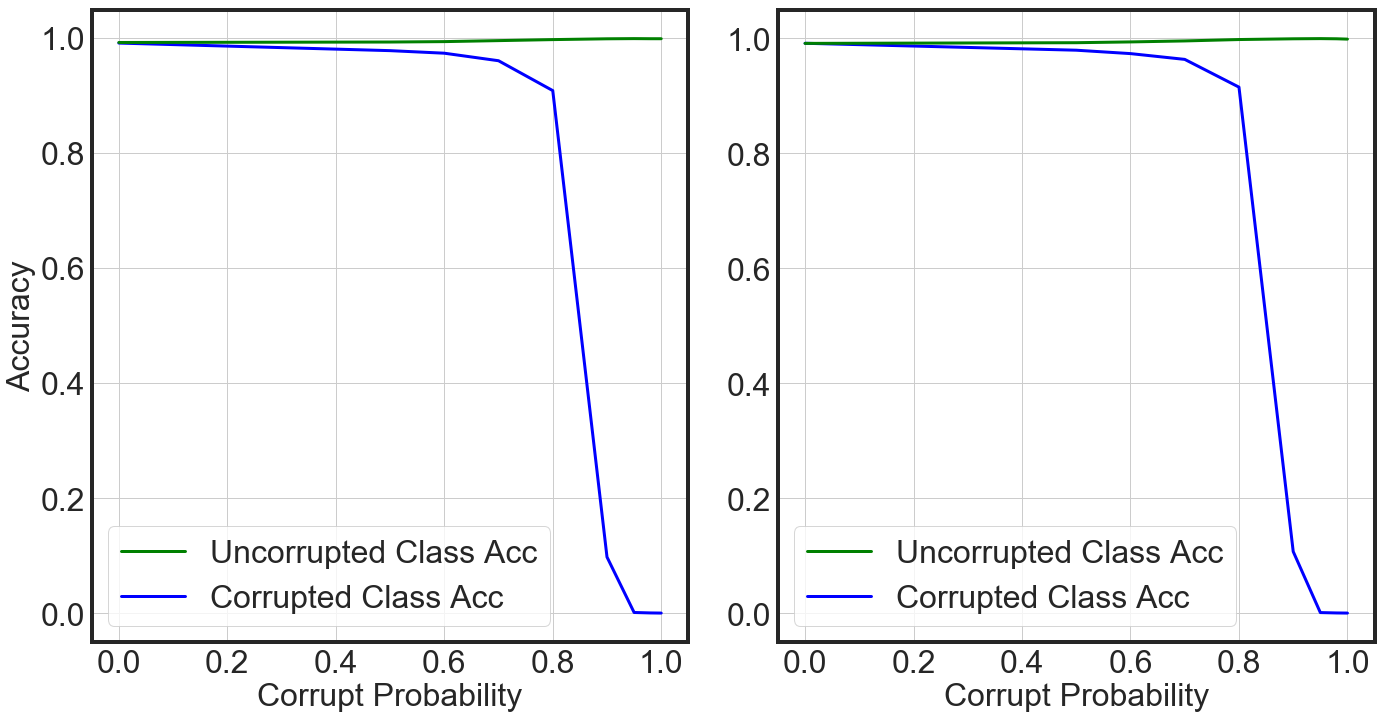}
    \label{fig:1}
    \caption{MNIST}
  \end{subfigure}
  \begin{subfigure}{0.49\textwidth}
    \includegraphics[width=1\linewidth]{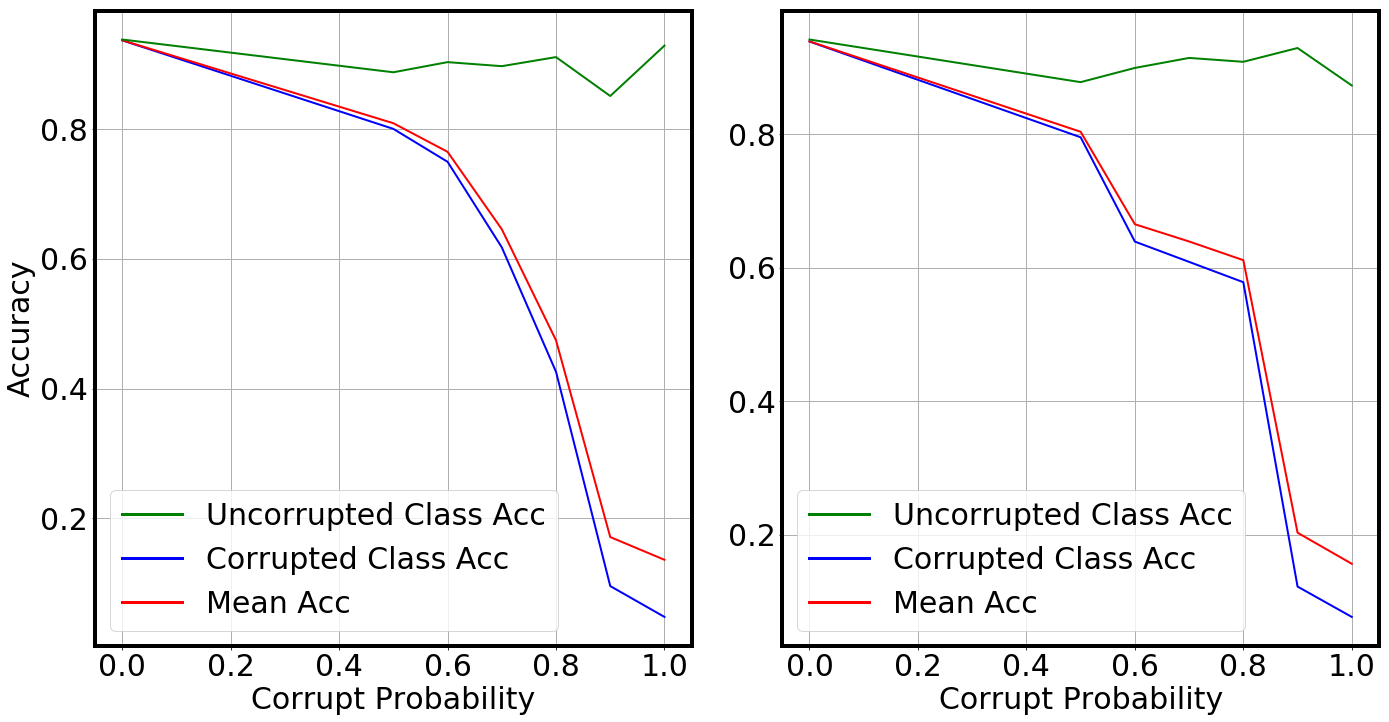}
    \label{fig:2}
    \caption{CIFAR10}
  \end{subfigure}
  \caption{Result of multi-class classification in MNIST and CIFAR10: Average classification accuracy in corrupted class and uncorrupted class as a function of corruption probability(strength). In the \textbf{left} plot of \textbf{(a)} and \textbf{(b)}, the corrupted label can take the label of uncorrupted class while in  the \textbf{right} plot of \textbf{(a)} and \textbf{(b)}, the corrupted label can not take the label of uncorrupted class. In both datasets, the classification accuracy of the uncorrupted class is not affected by the noisy labels of corrupted classes.
{\label{fig:mnistandcifar10}}%
}
\end{figure}
\subsection{Multi-Class Classification}

We present the classification accuracies of corrupted and uncorrupted classes as a function of corruption probability in figure \ref{fig:mnistandcifar10}. The trend lines start at zero corruption baseline and end horizontally at full corruption probability. 

As expected, the performance of corrupted classes in both MNIST and CIFAR10 datasets dropped with the increase in corruption probability as shown by figure \ref{fig:mnistandcifar10}. But, the performance drop was not seen until a large number of labels were corrupted, which highlights that neural networks are inherently robust to a certain level of noisy labels. In MNIST, the corrupted class performance didn't drop off until 80$\%$ corruption. Similar behavior is shown by the bigger network (ResNet18) in CIFAR10 dataset. The model's performance in corrupted classes started to drop off after 60$\%$ label corruption. We also saw that bigger networks start to memorize the noisy labels if overtrained. In contrast to LeNet-5, the ResNet18 architecture was able to overfit even the noisy labels at later epochs of training. This behavior is similar to what the \cite{Zhang_2021} paper discusses. These plots in figure \ref{fig:mnistandcifar10} also underline that the performance of the uncorrupted class remained almost consistent and didn't drop with an increase in the corruption probability. This behavior is consistent in both datasets. These experiments were done in two settings in which the corrupted label can either take the label of uncorrupted class or not. In both settings, we observed a similar trend in the accuracy of uncorrupted classes. The results strongly suggest that the class-dependent label noise of corrupted classes do not perturb the feature learning of the uncorrupted class.

\subsection{Multi-Task Learning}
As described in the experiment section, Task1 (50 classes of CIFAR100) benefits from training together with Task2 (other 50 classes of CIFAR100) in hard parameter sharing configuration. In single-task learning, the average accuracy of Task1 was about 75.63 $\%$. When Task1 was trained with Task2 in a multi-task learning setting, the average accuracy improved to 79.50 $\%$, which is about 3.87 raw improvement (equivalently 5.11 percentage improvement).

\begin{figure}[t]
\begin{center}
\includegraphics[width=0.6\columnwidth]{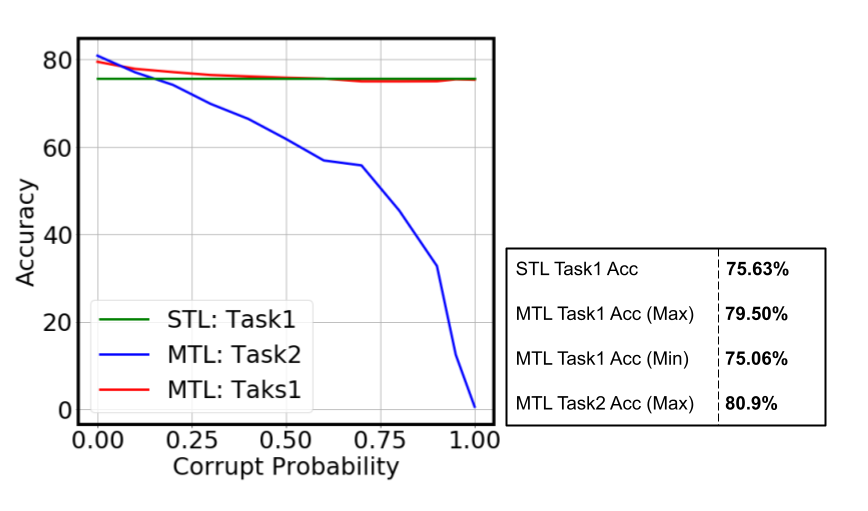}
\caption{{Average classification accuracy in two tasks as a function of label corruption in one of the tasks. Task1 represents the classification task of 50 classes of CIFAR100, while Task2 represents the classification task of the other 50 classes. Task2 labels are corrupted incrementally, while Task1 labels are not changed at all. STL (green line) is the single task learning baseline performance in Task1. MTL (blue line) and MTL (red line) are the performance of Task2 and Task1, respectively, in a multi-task learning setting. This plot shows the average value across three experiments (that used differently split 100 classes into two 50-50 class tasks) across each probability value.
{\label{fig:multitask}}%
}}
\end{center}
\end{figure}

The average classification accuracy in Task2 was 80.9$\%$ at zero corruption. As shown in Fig \ref{fig:multitask}, the blue trend line starts at high accuracy and falls with an increase in corruption probability. The line falls slowly at first but drops drastically later, approaching close to zero. The trend line of Task1 (red) is of greater importance to us than the trend in Task2. Initially, the line starts with a positive offset from the STL line due to the task transfer benefit achieved by training with Task2. Interestingly we start to lose the transfer benefit as Task2 is corrupted. At full corruption, the task transfer benefit completely vanishes and the multi-task learning performance in Task1 is almost the same as single-task performance. This result validates the hypothesis we proposed that the task-specific noisy labels impact the task transfer benefit. It is important to note that multi-task test performance does not fall far below the single task test performance line. It suggests that task-specific corruption doesn't introduce negative interference but only reduces the task transfer benefit achieved from multi-task learning.

\subsection{Multi-Label Learning}

\begin{figure}[t]
\begin{center}
\includegraphics[width=1\textwidth]{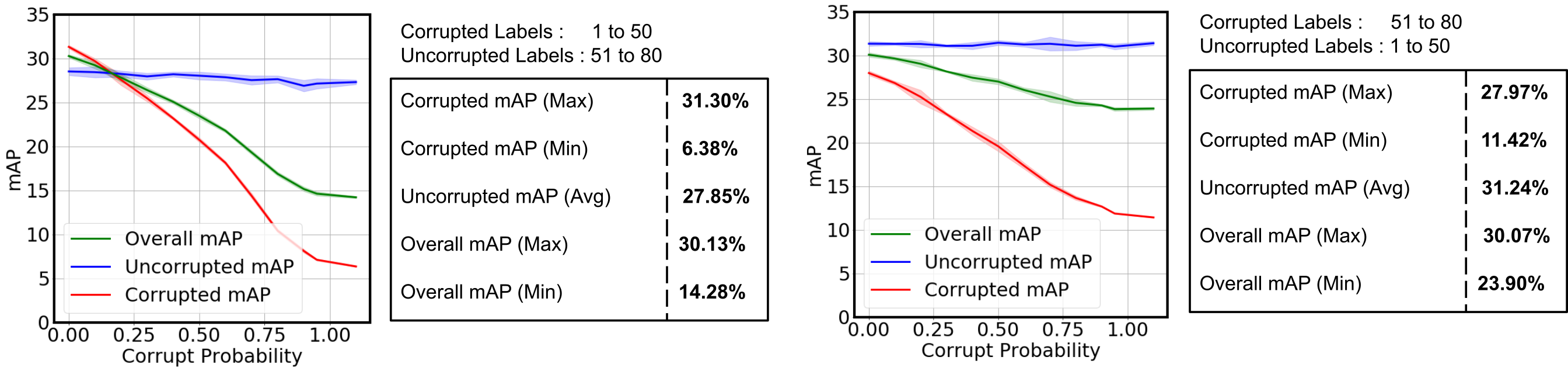}
\caption{{Results in Multi-Label Classification of mini MS-COCO dataset in two cases where the number of corrupted labels is either greater (\textbf{left} figure) or less (\textbf{right} figure) than the number of uncorrupted labels. Three trend lines: Uncorrupted mAP, Corrupt mAP, and Overall mAP, show the mean average precision as a function of corruption probability (strength) in labels belonging to the corrupted category. mAP of corrupted category and the overall dataset falls with increase in corruption probability. The consistent trend line of uncorrupted labels shows that their mAP isn't affected by label corruption in the other category. The shades in the line show the standard deviation across multiple experiments with random weight initialization. 
{\label{fig:multilabel}}
}}
\end{center}
\end{figure}

After the experiments in \ref{section:multi-label}, we analyzed the impact of label-dependent noisy labels in the performance of clean labels in the multi-label classification setting. Figure \ref{fig:multilabel} shows the mean average precision (mAP) as trend lines, which are functions of corruption probability in \textbf{category 1}. 

Our baseline, even at no corruption, starts with a lesser mAP than compared to the state-of-the-art methods for multi-label classification because we have used a smaller architecture ResNet18 in a naive configuration, without any hyperparameter tuning. State-of-the-art methods \cite{benbaruch2021asymmetric,zhu2017learning,ridnik2021imagenet21k} obtained higher mAP using larger model. We didn't use pre-trained weights as we are interested in finding if a network can generalize well from the start if we train with dependent heterogeneous noisy labels.
In the first case where the number of corrupted labels is greater than the uncorrupted labels, initially, the mAP in \textbf{category 1} is 31.30$\%$. With label corruption, the performance falls and ultimately drops to the minimum value of 6.38$\%$ at maximum corruption. A similar trend is seen in the second case, where the number of corrupted labels is less than the number of uncorrupted labels. In second case, the maximum mAP in \textbf{category 1} is 27.97$\%$ while the minimum mAP after full corruption is 11.42$\%$.

The trend line of the uncorrupted category (blue, \textbf{category 2}) shows that the corrupted labels didn't impact its performance at all. The line stays consistent throughout the horizontal axis. It is important to note that the corrupted mAP and uncorrupted mAP start at different offsets because the mean average precision across the two categories isn't the same. We also showed the overall mAP trend line follows the dominant category with more samples. Our work shows that in a naive setup, the label-dependent noisy labels do not affect the performance of clean labels even on training together. 

\section{Discussion}
Both the multi-class and the multi-label classification showed that the corrupted labels do not affect the network's performance in classifying clean labels. This behavior strongly hints that noisy labels don't impact the learning of low-level features much, otherwise, the network wouldn't have performed well in clean labels either. An interesting future work could be to investigate how the heterogeneous noisy labels impact the features learned at each layer in a deeper network.

In multi-task learning, we found that the performance in a clean task drops if the helping task is corrupted. While training, we also observed that the overfitting of label noise is an issue in a multi-task setting as well, which can be prevented by a properly tuned regularizer. But if the model somehow overfits, the performance in both the corrupted task and uncorrupted task drops no matter whether there is a positive or negative transfer between tasks. In such a case, we saw that the test performance in the clean task decreased below the single-task performance baseline, at high corruption in the helping task. However, if the overfitting was avoided, the corruption in helping task affected the clean task but the performance in the clean task didn't drop below its single task performance even at full corruption in helping task.

All our experiments investigated the classification case. However, in the real world, the heterogeneous label noise can appear in regression tasks as well. In future work, we are interested in investigating how the label noise in a certain range affects the overall generalization of the regression task.

\section{Conclusion}
In this work, we investigated the impact of heterogeneous noisy labels in three supervised classification settings. In multi-class classification, we found that training with the class-dependent noisy labels of certain classes doesn't affect the model's performance on classes with clean labels. On examining the multi-task learning, we observed that task transfer benefit is affected by task-dependent noise in helping tasks. The transferred benefit decreases with the increase in label corruption in helping tasks and reaches a minimum at full corruption where multi-task performance is approximately the same as single-task performance. Finally, we showed that label noise in certain categories of labels in the multi-label classification also doesn't affect the performance in clean labels when trained together.\\

\noindent\textbf{Acknowledgements.} This work was supported in part by the DARPA/SRI Lifelong Learning Machines program [HR0011-18-C-0051], AFOSR grant [FA9550-18-1-0121], and NSF award \#1909696. The views and conclusions contained herein are those of the authors and should not be interpreted as representing the official policies or endorsements of any sponsor.

\bibliography{mybib.bib}{}
\bibliographystyle{splncs04}

\end{document}